\documentclass{bmvc2k}
\usepackage{times}
\usepackage{epsfig}
\usepackage{graphicx}
\usepackage{amsmath}
\usepackage{amssymb}
\usepackage{bbm}
\usepackage{amsmath}
\usepackage{amssymb}
\usepackage{dsfont}
\usepackage{boldline}
\usepackage{booktabs}
\usepackage{gensymb}
\usepackage{times}  
\usepackage{helvet}  
\usepackage{courier}  
\usepackage{url}  
\usepackage{graphicx}  
\usepackage{multirow}
\usepackage{comment}
\usepackage{color}
\usepackage{etoolbox}
\usepackage{footnote}
\usepackage{multirow}
\usepackage{wrapfig,lipsum,booktabs}
\newcommand{\splitcell}[2][l]{\begin{tabular}[#1]{@{}l@{}}#2\end{tabular}}
\newcommand{\figref}[1]{Fig\onedot~\ref{#1}}

\newcommand{\secref}[1]{Sec\onedot~\ref{#1}}
\newcommand{\tabref}[1]{Tab\onedot~\ref{#1}}

\newcommand{\ve}[1]{{\mathbf #1}} 
\newcommand{\hua}[1]{{\mathcal #1}}

\newcommand{\thickhline}{%
    \noalign {\ifnum 0=`}\fi \hrule height 1pt
    \futurelet \reserved@a \@xhline
}

\DeclareRobustCommand\onedot{\futurelet\@let@token\@onedot}
\def\onedot{\ifx\@let@token.\else.\null\fi\xspace}
\def\eg{\emph{e.g.}} 

\def\any{\forall}
\def\ie{\emph{i.e.}}

\def\wrt{w.r.t\onedot}

\title{EPNAS: Efficient Progressive Neural Architecture Search}

\addauthor{Yanqi Zhou$^*$}{yanqiz@google.com}{1}
\addauthor{Peng Wang}{wangpeng54@baidu.com}{3}
\addauthor{Sercan Arik$^*$}{soarik@google.com}{2}
\addauthor{Haonan Yu$^*$}{haonanu@gmail.com}{4}
\addauthor{Syed Zawad$^*$}{szawad@nevada.unr.edu}{5}
\addauthor{Feng Yan}{fyan@unr.edu}{5}
\addauthor{Greg Diamos$^*$}{greg@landing.ai}{6}

\addinstitution{
 Google Brain
}
\addinstitution{
 Google Cloud AI
}
\addinstitution{
 Baidu Research
}
\addinstitution{
 Horizon Robotics
}
\addinstitution{
 University of Nevada
}
\addinstitution{
 Landing AI
}
\runninghead{Yanqi Zhou et.al.}{EPNAS}

\def\eg{\emph{e.g}\bmvaOneDot}

\begin{document}

\maketitle
\begin{abstract}
In this paper, we propose Efficient Progressive Neural Architecture Search (EPNAS), a neural architecture search (NAS) that efficiently handles large search space through a novel progressive search policy with performance prediction based on REINFORCE~\cite{Williams.1992.PG}. EPNAS is designed to search target networks in parallel, which is more scalable on parallel systems such as GPU/TPU clusters. 
More importantly, EPNAS can be generalized to architecture search with multiple resource constraints, \eg, model size, compute complexity or intensity, which is crucial for deployment in widespread platforms such as mobile and cloud.  
We compare EPNAS against other state-of-the-art (SoTA)  network architectures (\eg, MobileNetV2~\cite{mobilenetv2}) and efficient NAS algorithms (\eg, ENAS~\cite{pham2018efficient}, and PNAS~\cite{Liu2017b}) on image recognition tasks using  CIFAR10 and ImageNet. On both datasets, EPNAS
is superior \wrt architecture searching speed and recognition accuracy. 

\end{abstract}
\vspace{-0.5\baselineskip}
\section{Introduction}
\label{sec:intro}
\noindent Deep neural networks have demonstrated excellent performance on challenging tasks and pushed the frontiers of impactful applications such as image recognition \cite{imagerecognition1}, image synthesis \cite{imagegeneration1}, language translation \cite{MT1}, speech recognition and synthesis \cite{DS2, parallelwavenet}. Despite all these advancements, designing neural networks remains as a laborious task, requiring extensive domain expertise. Motivated by automating the neural network design while achieving superior performance, neural architecture search (NAS) has been proposed \cite{real2017large,Zoph2016,Liu2017b,Baker2016,Negrinho2017}.

Conventional NAS algorithms are performed with limited search spaces (\eg~small number of operation type) due to lack of efficiency, which hinders the application of NAS to various tasks. 
For example, ~\cite{Zoph2016} uses 800 GPUs, and takes 3 days to discover a model on a small dataset like CIFAR10~\cite{krizhevsky2014cifar}, which is infeasible to directly search models over larger datasets such as COCO~\cite{lin2014microsoft} or ImageNet~\cite{deng2009imagenet}. 
Therefore, it is crucial to improve searching efficiency of NAS to allow larger space of architecture variation, in order to achieve better model performance or handle multiple objective simultaneously.

Commonly, the efficiency of a NAS algorithm depends on two factors: 1) the number of models need to be sampled. 2) The time cost of training and evaluating a sampled model.
Targeting at the two aspects, recent works~\cite{pham2018efficient,liu2018darts} have made significant progress.
For example, ENAS~\cite{pham2018efficient},  DARTS~\cite{liu2018darts} or SNAS \cite{xie2019snas} try to share the architecture and parameters between different models trained during searching period, thus reduce the time cost of model training. However, their models are searched within a large pre-defined graph that need to be fully loaded into memory, therefore the search space is limited.
EAS~\cite{Han.2017.NT} and PNAS \cite{Liu2017b} progressively add or morph an operation based on a performance predictor or a policy network, thus can support larger operation set at search time and significantly reduce the number of models need to be sampled. 
This paper further explores the direction of progressive search, aiming to improve sample efficiency and generality of search space. We call our system efficient progressive NAS (EPNAS). Specifically, based on the framework of REINFORCE~\cite{Williams.1992.PG}, to reduce necessary number of sampled models, we design a novel set of actions for network morphing such as group-scaling and removing, additional to the adding or widening operations as proposed in EAS~\cite{Han.2017.NT}. It allows us to initiate NAS from a better standing point, \eg~a large random generated network~\cite{xie2019exploring}, rather than from scratch as in PNAS or a small network as in EAS. 
Secondly, to reduce the model training time, we propose a strategy of aggressive learning rate scheduling and a general dictionary-based parameter sharing, where a model can be trained with one fifth of time cost than training it from scratch. 
Comparing to EAS or PNAS, as shown in~\secref{sec:exp}, EPNAS provides 2$\times$ to 20$\times$ overall searching speedup with much larger search space, while providing better performance with the popular CIFAR10 datasets. Our model also generalizes well to larger datasets such as ImageNet. 

In addition to model accuracy, deep neural networks are deployed in a wider selection of platforms (e.g., mobile device and cloud) today, which makes resource-aware architecture design a crucial problem. 
Recent works such as MNASNet~\cite{tan2018mnasnet} or DPPNet~\cite{dong2018dpp} extend NAS to be device-aware by designing a device-aware reward during the search. In our case, thanks to the proposed efficient searching strategy, EPNAS can also be easily generalized to multi-objective NAS which jointly considers computing resource related metrics, such as the model's memory requirement, computational complexity, and power consumption. Specifically, we transform those hard computational constraints to soft-relaxed rewards for effectively learning the network generator. In our experiments, we demonstrate that EPNAS is able to perform effectively with various resource constrains which fails random search. 

Finally, to better align NAS with the size of datasets, we also supports two search patterns for EPNAS. For handling small dataset, we use an efficient layer-by-layer search that exhaustively modify all layers of the network. For handling larger dataset like ImageNet, similar with PNAS, we adopt a module-based search that finds a cell using small dataset and stack it for generalization.
We evaluate EPNAS's performance for image recognition on CIFAR10 and its generalization to ImageNet. 
For CIFAR10, EPNAS achieves 2.79\% test error when compute intensity is greater than 100 FLOPs/byte, and 2.84\% test error when model size is less than 4M parameters. For ImageNet, we achieve 75.2\% top-1 accuracy with 4.2M parameters. In both cases, our results outperform other related NAS algorithms such as PNAS~\cite{Liu2017b}, and ENAS~\cite{Pham2018}.

\vspace{-1\baselineskip}
\section{Related work}
\vspace{-0.5\baselineskip}


\vspace{-0.5\baselineskip}
\paragraph{Neural architecture search (NAS).}
NAS has attracted growing interests in recent years due to its potential advantages over manually-crafted architectures. As summarized by~\cite{elsken2018neural}, the core issues lie in three aspects: efficient search strategy, large search space, and integration of performance estimation. 
Conventionally, evolutionary algorithms \cite{angeline1994evolutionary,stanley2002evolving,Liu2017HR,real2017large,Esteban.2018.REIC,Liu2017b} are one set of methods used for automatic NAS. NAS has also been studied in the context of Bayesian optimization \cite{Bergstra.2012.RSH,KK.2018.NASBOO}, which models the distribution of architectures with Gaussian process.
Recently, reinforcement learning \cite{Baker2016,Zoph2016,Baker} with deep networks has emerged as a more effective method. However, these NAS approaches are computationally expensive, with relatively small search space \wrt single target of model accuracy, e.g., \cite{Zoph2016} used hundreds of GPUs for delivering a model comparable to human-crafted network on CIFAR10.
To tackle the search cost, network morphism \cite{Elsken.2017.MOAS} for evolutionary algorithm, and efficient NAS approaches with parameter sharing such as ENAS \cite{pham2018efficient} are proposed. 
Specifically, 
ENAS employs weight-sharing among child models to eschew training each from scratch until convergence. To handle larger datasets, one-shot NAS such as path selection~\cite{bender2018understanding} and operation selection such as DARTS~\cite{liu2018darts} are also proposed.
However, parameter sharing as ENAS and operation sharing as DARTS search within a pre-defined network graph, which limits the scaling of search space, \eg, channel size and kernel size cannot be flexibly changed for each layer in order to reuse the operations or weights. 
However, \textbf{larger search space is crucial for discovering architectures especially when multiple objective or resource constraints are also considered}.
Another set of methods for reducing search cost is to progressively adapt a given architectures such as EAS~\cite{Han.2017.NT} and PNAS~\cite{Liu2017b}. In each step, one may freely choose to add an operation from a set, which enables larger searching possibility for networks or a cell structure.
Therefore, we follow this learning strategy and propose a novel progressive search policy with performance prediction, which brings additional efficiency and can reach higher accuracy as demonstrated in our experiments. 

\paragraph{Resource-constraint NAS.} 
\vspace{-0.5\baselineskip}
It used to be that most effective approaches for optimizing performance under various resource constraints rely on the creativity of the researchers. Among many, some notable ones include attention mechanisms \cite{showattentell}, depthwise-separable convolutions \cite{depthwisesepconv,MobileNet}, inverted residuals \cite{mobilenetv2}, shuffle features~\cite{ZhangZLS17shuffle, ma2018shufflenet}, and structured transforms \cite{structuredtransforms}. 
There are common approaches that reduce model size or improve inference time as a post processing stage.
For example, sparsity regularization \cite{sparsity1}, channel pruning~\cite{He_2017_ICCV}, connection pruning \cite{pruning1}, and weights/activation quantization \cite{quantization1} are common model compression approaches. 
Similarly, one may automatically design a resource constraint architecture through NAS. Most recently, AMC~\cite{he2018amc} adopts reinforcement learning to design policy of compression operations. 
MorphNet~\cite{gordon2018morphnet} proposes to morph an existing network, \eg~ResNet-101, by changing feature channel or output size based on a specified constraints. DPPNet~\cite{dong2018dpp} or MNASNet~\cite{tan2018mnasnet} propose to directly search a model with resource-aware operations based on PNAS~\cite{liu2018darts} or NAS~\cite{Zoph2017} framework respectively.
In our work, as demonstrated in our experiments, EPNAS also effectively enables NAS with multiple resource requirements simultaneously thanks to our relaxed objective and efficient searching strategy.

\vspace{-1\baselineskip}
\section{Efficient progressive NAS with REINFORCE}
\vspace{-0.5\baselineskip}
\label{sec:progressive}
In this section, for generality, we direct formulate EPNAS with various model constraints, as one may simply remove the constraints when they are not necessary. Then, we elaborate our optimization and  proposed architecture transforming policy networks.

As stated in~\secref{sec:intro}, rather than rebuilding the entire network from scratch, we adopt a progressive strategy with REINFORCE~\cite{Williams.1992.PG} for more efficient architecture search so that architectures searched in previous step can be reused in subsequent steps. 
Similar ideas are proposed for evolutionary algorithms~\cite{Liu2017b} recently, however, in conjunction with reinforcement learning (RL), our method is more sample efficient.
This strategy can also be considered in the context of Markov Chain Monte Carlo sampling~\cite{brooks2011handbook} (as the rewards represent a target distribution and the policy network approximates the conditional probability to be sampled from), which has been proven to be effective in reducing the sampling variation when dealing with high dimensional objective, yielding more stable learning of policy networks. 

Formally, given an existing network architecture $\hua{X}$, a policy network $\pi_\theta(\ve{a}|\hua{X})$ generates action $\ve{a}$ that progressively change $\hua{X}$ from our search space $\hua{S}$, \wrt 
multiple resource constraints objective:
\begin{equation}
\begin{split}
&~~\max\nolimits_{\hua{X} \in \hua{S}} P(\hua{X} | \hua{D}) \\
&~~s.t.~\mathds{1}(U_i(\hua{X}) \in [C_{li}, C_{ui}]), ~~\any i \in \{1, \dots, K\},
\end{split}
\label{eqn:comb}
\end{equation}
where $P(\hua{X} | \hua{D})$ is the performance of $\hua{X}$ under dataset $\hua{D}$.  $\mathds{1}()$ is an indicator function, and $U_i(\hua{X})$ is the $i_{th}$ resource usage of the network, $C_{li}$ and $C_{ui}$ are the corresponding constraint, indicating the lower bound and upper bound for resource usage. Next, we show how to optimize the objective with policy gradient for progressive architecture generation. 

\vspace{-1\baselineskip}
\subsection{Policy Gradient with Resource-aware Reward}
\begin{figure}[t]
\centering
\includegraphics[width=0.6\linewidth,trim={2cm 3cm 2cm 3cm},clip]{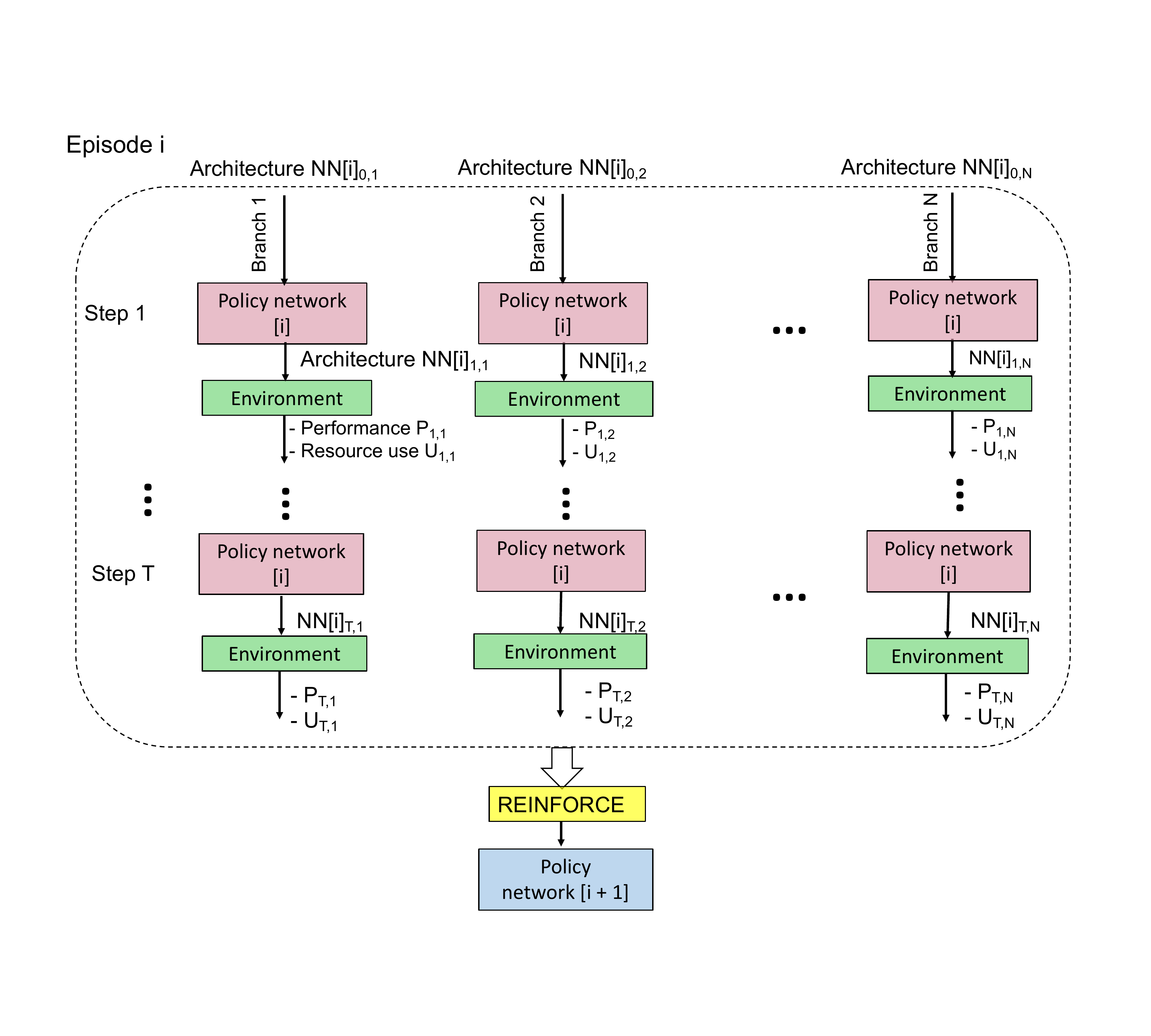}
\vspace{-1\baselineskip}
\caption{REINFORCE step for policy gradient. $N$ is the number of parallel policy networks to adapt a baseline architecture at episode of $i$.}
\label{fig:rl_algo}
\vspace{-1\baselineskip}
\end{figure}

Given the objective in Eqn.~(\ref{eqn:comb}), we optimize it following the procedure to train a policy network $\pi_\theta(\cdot)$ using policy gradient \cite{Williams.1992.PG}, by taking the reward $r(\hua{X})$ as Lagrange function of Eqn.~(\ref{eqn:comb}).
We depict the optimization procedure in Fig.~\ref{fig:rl_algo}, where the policy networks are distributed to $N$ branches. For a branch $n$, at episode $e$, we manipulate the network from an initial network $\hua{X}_{0,n}$, and at step $t$, the policy network generates the action $\ve{a}_{t,n}$, which transforms the network $\hua{X}_{t,n}$ to $\hua{X}_{t+1,n}$ that is evaluated using the reward. We adopt $T$ steps for an episode, and accumulate all the rewards over all steps and all batches to compute the gradients. Formally, the  gradient for parameter $\theta$ is
\begin{equation}
g = \frac{1}{N}\sum_{n=0}^{N-1}\sum_{t=0}^{T-1}
\bigtriangledown_{\theta}
\log\pi_{\theta}(\ve{a}_{t,n}|\hua{X}_{t,n})
\left (\sum_{t^{'}=t}^{T-1}r_{t^{'},n}-b(
\hua{X}_{t}) \right ). \nonumber
\end{equation}
where $\sum_{t^{'}=t}^{T-1}r_{t^{'},n}-b(\hua{X}_{t})$ is used as an approximate estimate of the efficacy of the action $\ve{a}_{t,n}$.
Ideally, we may train neural networks to convergence until the objective is fully optimized and all constraints are satisfied. However, empirically, hard constraints such as an indicator function in Eqn.~(\ref{eqn:comb}) yields extremely sparse reward. We can not obtain any feedback if one constraint is violated. 
Thus, we relax the hard constraint from an indicator function to be a soft one using a hinge function, where models violating the constraint less are rewarded higher. Formally, we switch the binary constraint in Eqn.~(\ref{eqn:comb}) to be a soft violation function: 
\begin{equation}
\hua{V}(U_i)=p_i^{\max(\max(0,  U_i/C_{ui} - 1), \max(0,  C_{li}/U_i-1))},
\end{equation}
where $p_i \in [0,1]$ is a base penalty hyperparameter used to control the strength of the soft penalty in terms of the degree of violation \wrt~$i_{th}$ constraint. 
\footnote{Each constraint may use a different base penalty hyperparameter in principle, but we simply use $p_i$ = 0.9 for all architecture search experiments and show that it is effective to constrain the search for all constraints.}. Finally, we switch our reward function navigating the search in the constrained space as,
\begin{equation}
\label{reward}
    r_{t,n} = P_{t,n}\prod_{i=1}^{K}\hua{V}(U_i).
\end{equation}
where $K$ is the number of constraints. We train the policy network till convergence. Empirically, EPNAS always finds a set of models satisfying our original objective. We then select the best one, yielding a resource-constraint high-performance model. 
Later, we will introduce the designed policy network and the action space for progressive network generation.



\vspace{-0.5\baselineskip}
\subsection{Policy Network}
\label{sec:policy}
\vspace{-0.3\baselineskip}

\begin{figure}[!tp]
\centering
\includegraphics[width=0.6\linewidth]{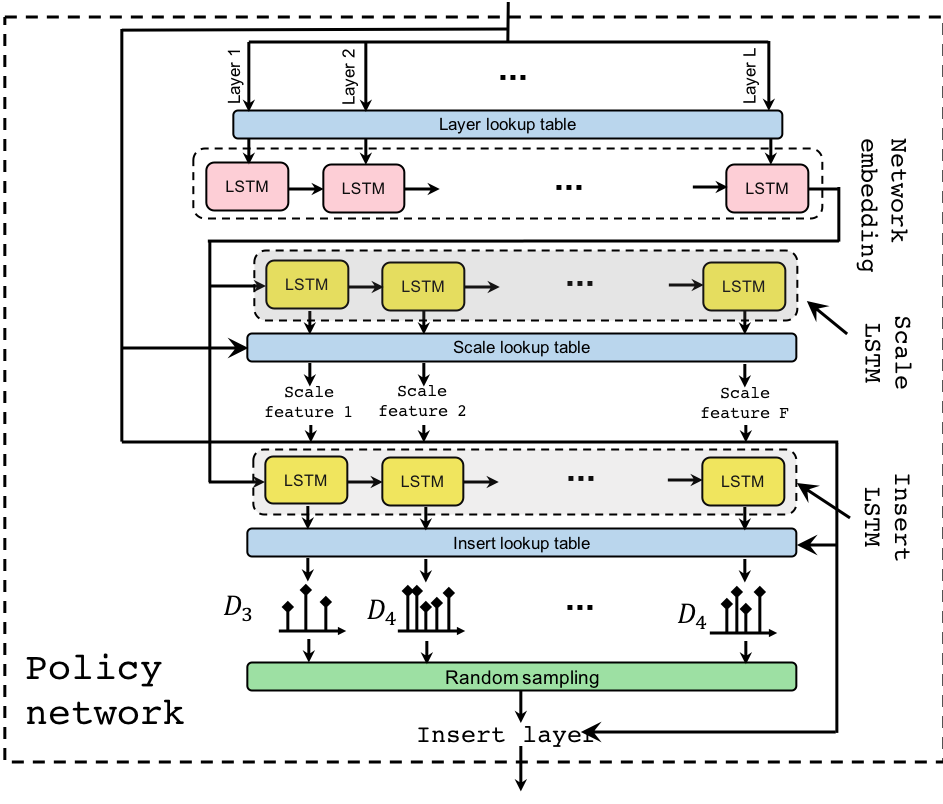}
\vspace{-1\baselineskip}
\caption{Policy Network of EPNAS. It is an LSTM-based network, which first generates network embedding, and then outputs actions to modify the network with a ``Scale LSTM'' and an ``Insert LSTM''. Details in \secref{sec:policy}. 
}
\vspace{-0.5\baselineskip}
\label{fig:policy-network-with-embedding}
\end{figure}
Policy network, shown in Fig.~\ref{fig:policy-network-with-embedding}, adapts any input network by progressively modifying its parameters (referred as the \emph{scale action}), or by inserting/removing a layer (referred jointly as the \emph{insert action}). 
At every training step $t$, rather than building the target network from scratch~\cite{Zoph2017,Pham2018}, EPNAS modifies the architecture from preceding training step via previously described operations. This progressive search enables a more sample-efficient search.

We use a \emph{network embedding} to represent the input neural network configuration. Initially, each layer of the input neural network is mapped to layer embeddings by a trainable embedding lookup table.
Then, an LSTM layer (with a state size equal to the number of layers $L$) sequentially processes these layer embeddings and outputs a network embedding. Next, the network embedding is input to two different LSTMs that decide the scale, insert, and remove actions. 
The first LSTM, named as ``scale LSTM'', outputs the hidden states at every step $t$ that correspond to a scaling action $\ve{a}_s$ from a scale lookup table which changes the filter width or change the number of filters, for the network. 
For example, in layer-by-layer search, we may partition the network into $F$ parts, and for part $f$, the output $\ve{a}_{sf}$ can scale a group of filters simultaneously. In our experiments, this is more efficient than per-layer modification of ENAS~\cite{Pham2018}, while more flexible than global modification of EAS~\cite{Han.2017.NT}. 
The second LSTM, named as ``insert LSTM'', outputs hidden states representing actions $\ve{a}_i$, which selects to do insert, remove, or keep a certain layer, \eg, using a \textit{conv} operation from the insert lookup table, at certain place $l$ after the network is scaled. 
To encourage exploration of insert LSTM, \ie, encourage inserting inside the network to explore inner structure, rather than always appending at the end of the network, we constraint the output distribution of inserting place $l$ to be determinant on layer number of input network $L$ using a check table. Formally, we let $\ve{P}(\ve{a}_i(p)=l)\sim \hua{D}_L$, where $\hua{D}_L$ stands for a discrete distribution with $L+5$ values. We use this prior knowledge, and found the architectures are more dynamically changed during search and able to find target model more efficiently.

\subsection{Search patterns}
\vspace{-0.5\baselineskip}
Following common strategies~\cite{Pham2018,Liu2017b}, EPNAS also supports two types of search patterns: (i) layer-by-layer search, which exhaustively modifies each layer of a network when dataset is not large, and (ii) module search, which searches a neural cell that can be arbitrarily stacked for generalization. The former is relatively slow, while can reach high performance, and the latter targets at architecture transformation applied on large datasets.
We elaborate the two search patterns in this section.

\begin{figure}[t]
    \centering
    \includegraphics[width=0.48\linewidth,trim={2cm 0.5cm 8cm 2cm},clip]{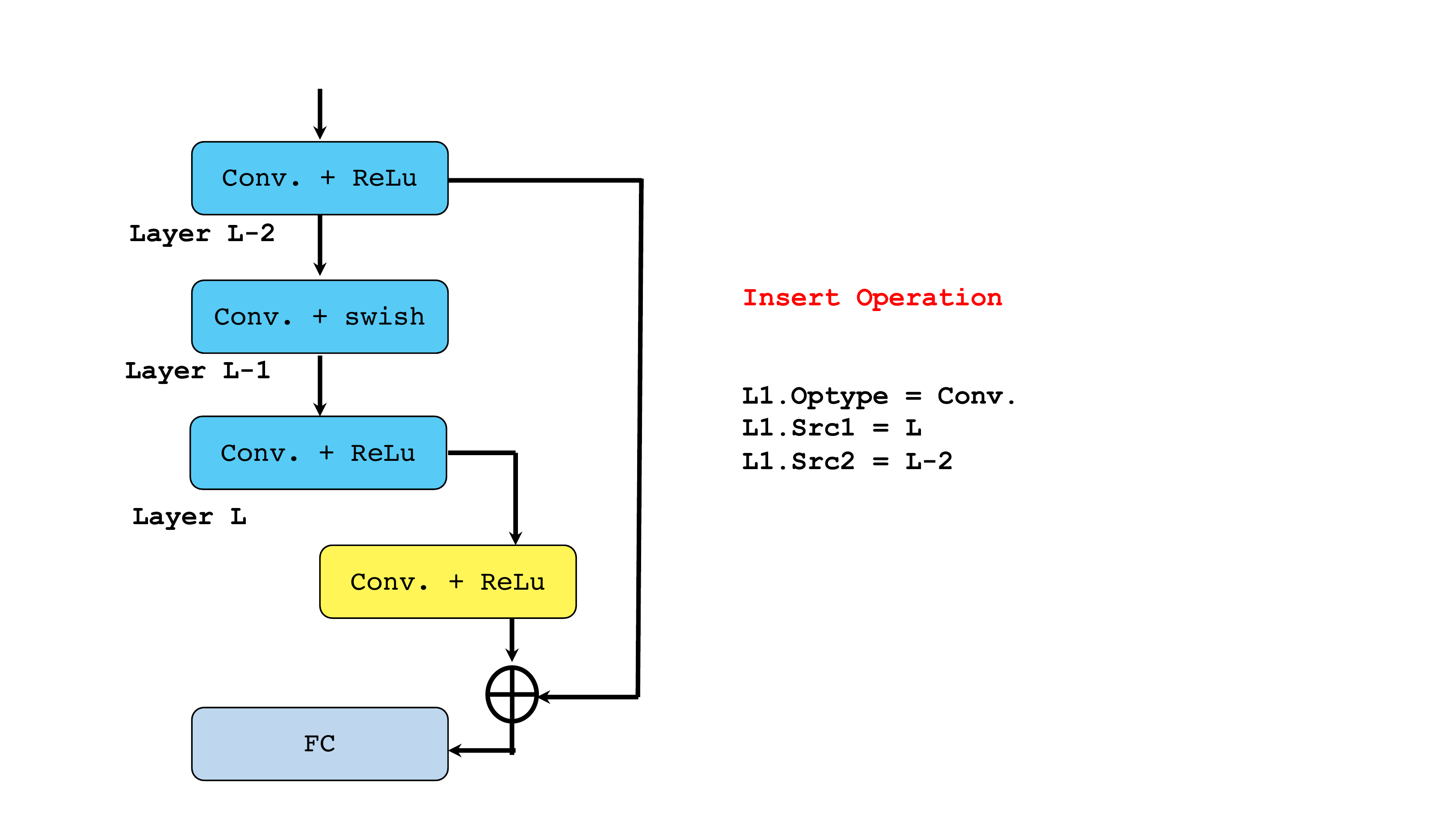}
    \includegraphics[width=0.48\linewidth,trim={1cm 4.5cm 1cm 3cm},clip]{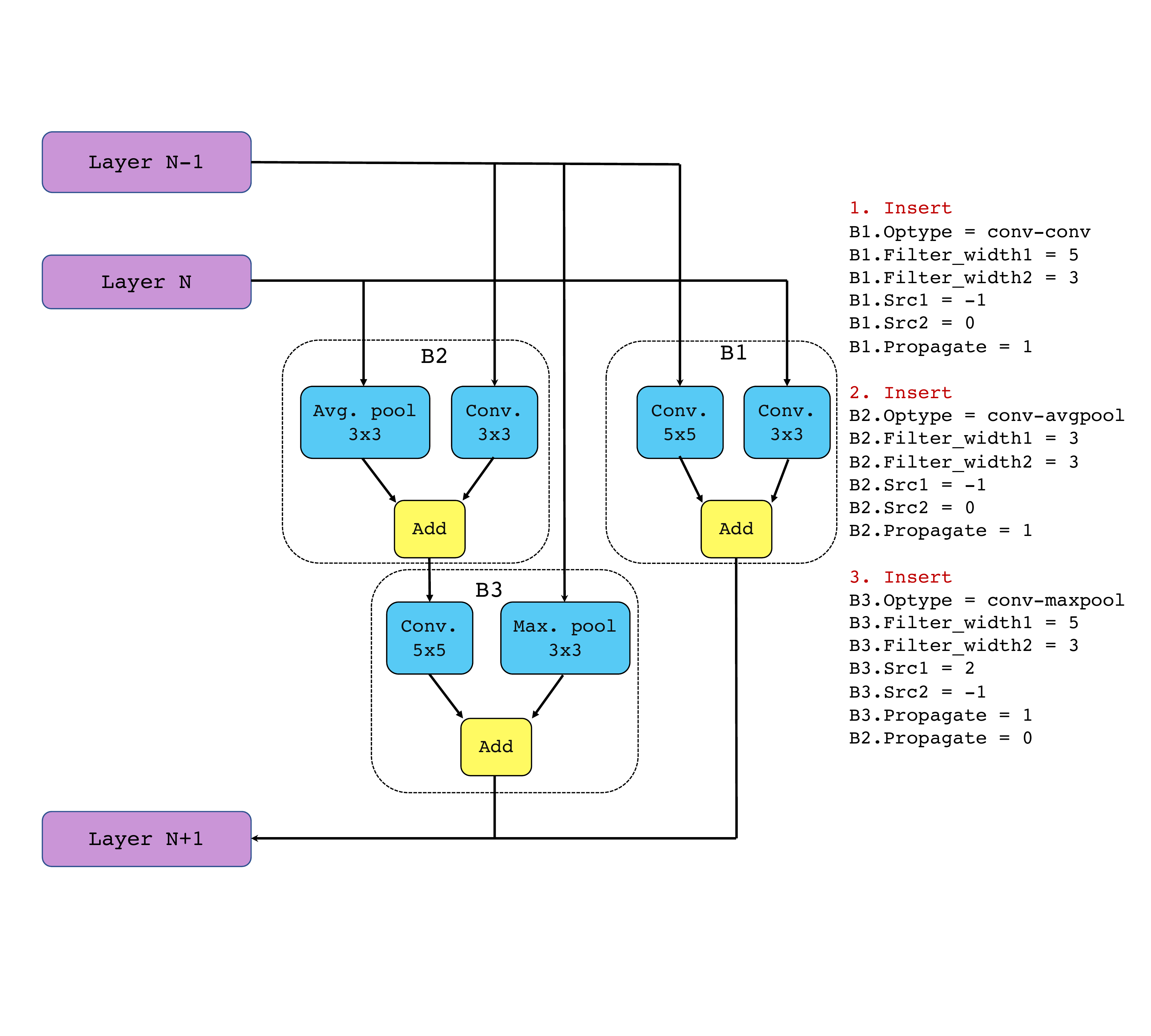}
    \vspace{-0.5\baselineskip}
    \caption{Left: A layer-by-layer search insert operation example. A conv operation is inserted after layer L, and has skip connection with layer L-2. Right: A module search insert operation example. When the branch 3 is inserted, one of its source value is from branch 2. After insertion, the connection between branch 2 and the next layer is cut off.}
    \label{fig:two-ops}
    \vspace{-1\baselineskip}
\end{figure}

In {\bf Layer-by-layer search},
EPNAS progressively scales and inserts layers that are potentially with skip connection. 
Fig.~\ref{fig:two-ops} exemplifies an insert operation by the ``Insert LSTM'', which is applied to an input network. ``Src1'' determines the place to insert a given layer (a conv layer), and ``Src2'' indicates where to add a skip connection. Here, ``Src2'' can be \textit{-1} to avoid a skip connection.
Specifically, the search operations are chosen based on the problem domain. For example, one can include recurrent layers for discriminative speech models and convolution layers for discriminative image models. More details are explained in Section~\ref{sec:exp}.

{\bf Module search} aims to find an optimal small network cell which can be repeatedly stacked to create the overall neural network. 
It has relatively limited search space with multi-branch networks. 
The insert action by the ``Insert LSTM'' no longer inserts a layer, but inserts a ``branch'', and outputs the types of the operation and corresponding scaling parameters (filter width, pooling width, channel size, etc.). 
Similar to layer-by-layer search, each branch consists of two operations to be concatenated. We illustrate an insert operation in Fig.~\ref{fig:two-ops}. Here, ``Src1'' and ``Src2'' determine where these two operations get input values from, and ``propagate'' determines whether the output of the branch gets passed to the next layer.  


\vspace{-0.5\baselineskip}
\subsection{Speedup EPNAS with performance prediction}
\vspace{-0.5\baselineskip}
Training every sampled model till convergence can be time-consuming and redundant for finding relative performance ranking. 
We expedite EPNAS with a performance prediction strategy that provides efficient training and evaluation.
Our approach is based on the observation that a similar learning pattern can be maintained as long as the learning rate and batch size are kept proportionally ~\cite{increasebatchsize2018}. By scaling up the batch size, we can use a large learning rate with an aggressive decay, which enables accurate ranking with much fewer training epochs. Besides, since relative performance ranking is more important, learning rate and batch size ratio can be increased. 
Additionally, we also allow the generated networks to partially share parameters across steps. Differently, in our cases, we have additional operations rather than just insertion and widening in EAS. 
Therefore, we use a global dictionary that is shared by all models within a batch of searched models. The parameters are stored as key-value pairs, where the keys are the combination of the layer number, operation type, and the values are corresponding parameters. At the end of each step, if there is a clash between keys, the variables for the model with the highest accuracy is stored.  
For operations where the layer and operation types match, but the dimensions do not (e.g., after scaling a layer), the parameters with the closest dimensions are chosen and the variables are spliced or padded accordingly. 
Equipped with performance prediction, EPNAS can find the good models within much less time cost,  which is critical for NAS with multiple resource constraints.
\begin{table*}[t]
\caption{Search space of scale and insert actions}
\begin{minipage}{.45\linewidth}
	\centering
    \caption{Layer-by-layer search.}
    \fontsize{6}{7}\selectfont
    \begin{tabular}{>{\centering\arraybackslash}p{1.9 cm}|>{\centering\arraybackslash}p{2 cm}}
 		\toprule[0.1 em]
		Feature &  Search space \\ \hline\hline
		Layer type & [conv2d, dep-sep-conv2d, MaxPool2d, AvgPool2d, add] \\\hline
	    Filter width & [1, 3, 5, 7] \\ \hline
	    Pooling width & [2, 3] \\ \hline
	    Channel size & [16, 32, 64, 96, 128, 256] \\\hline
	    Nonlinear activation & ["relu", "crelu", "elu", "selu", "swish"] \\\hline
	    Src1 Layer & [i for i in range(MAX\_LAYERS)] \\\hline
	    Src2 Layer & [i for i in range(MAX\_LAYERS)] \\\toprule[0.1 em]
   \end{tabular}
\end{minipage}%
\begin{minipage}{.45\linewidth}
	\centering
    \caption{Module search.}
    \fontsize{6}{8}\selectfont
	\begin{tabular}{>{\centering\arraybackslash}p{2.2 cm}|>{\centering\arraybackslash}p{3 cm}}
 		\toprule[0.1 em]
		Feature &  Search space \\ \hline\hline
		Branch type & [conv-conv, conv-maxpool, conv-avgpool, conv-none, maxpool-none, avgpool-none,$1\times7$-$7\times1$-none] \\\hline
	    Filter width & [1, 3, 5, 7] \\ \hline
	    Pooling width & [2, 3] \\ \hline
	    Channel size & [8, 12, 16, 24, 32] \\\hline
	    Src1 Layer & [i for i in range(MAX\_BRANCHES+1)] \\\hline
	    Src2 Layer & [i for i in range(MAX\_BRANCHES+1)] \\\hline
	    Propagate & [0,1] \\\toprule[0.1 em]
		\end{tabular}
    \end{minipage}
    \label{tab:insert_space_image}
\end{table*}

\begin{figure}[!htpb]
    \centering
    \includegraphics[width=0.8\linewidth]{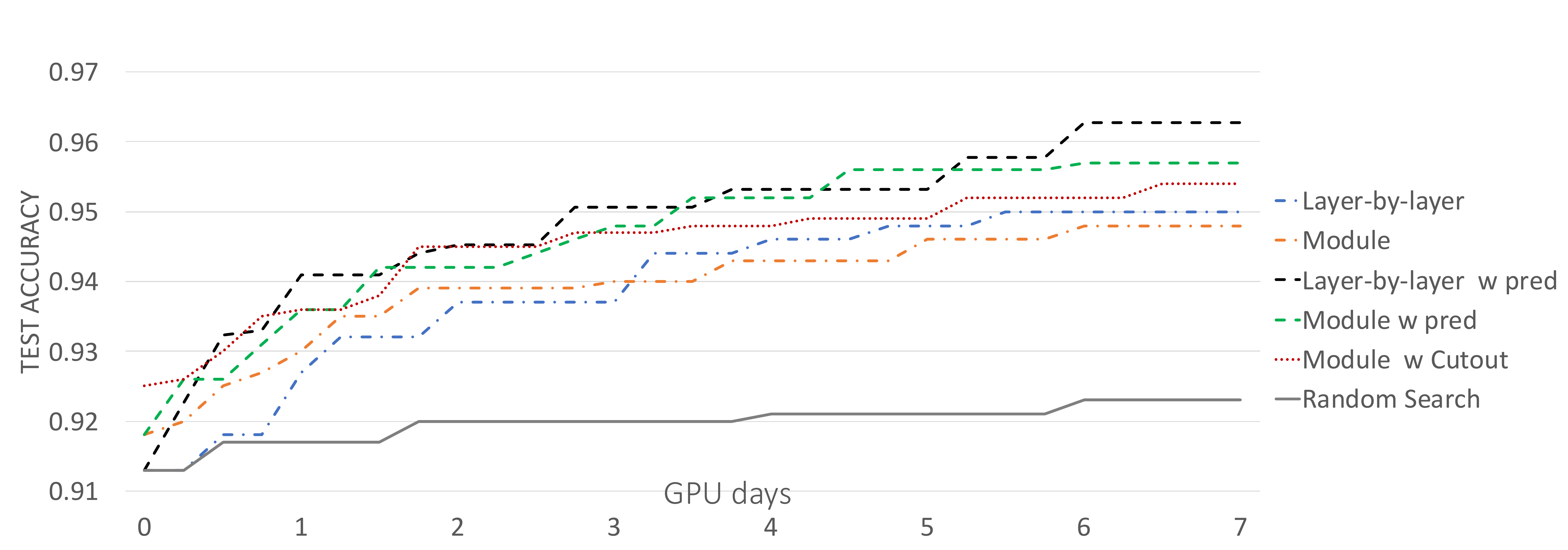}
    \caption{Accuracy VS. total search time for CIFAR-10. Note the accuracy reported here is only accuracy during searching. The final reported results in ~\tabref{tab:comparison_SOTA} and ~\tabref{tab:comparison_constrained} are evaluated by training from scratch.}
    \label{fig:na-accuracy}
    \vspace{-2mm}
    \vspace{-1\baselineskip}
\end{figure}

\vspace{-1\baselineskip}
\section{Experimental Evaluation}
\vspace{-0.5\baselineskip}
\label{sec:exp}
In this section, we evaluate the efficiency of EPNAS for handling large search space over two popular datasets:  CIFAR-10 dataset~\cite{krizhevsky2009learning} and ImageNet~\cite{deng2009imagenet}.  We first show model accuracy versus search time for five different search approaches used by EPNAS in~\figref{fig:na-accuracy}. Then, we compare EPNAS with other related NAS algorithms in ~\tabref{tab:comparison_SOTA} and show its generalization to ImageNet. Finally, we compare EPNAS under three resource constraints, \ie~model size ($sz$), compute complexity ($c.c.$) and compute intensity ($c.i.$), with SoTA models in ~\tabref{tab:comparison_constrained}. 
Due to space limitation, we put the details of resource constraints, visualization of searched models for layer-by-layer and module search in our supplementary materials. In our supplementary files, we also experimented our algorithm to a language dataset, \ie~keyword spotting using Google Speech Commands dataset~\cite{KWS-data}, and obtain SoTA results.

\vspace{-0.5\baselineskip}
\subsection{Experiments with CIFAR10}
CIFAR-10 dataset contains 45K training images and 5K testing images with size of $32\times 32$.
Following PNAS~\cite{Liu2017b}, we apply standard image augmentation techniques, including random flipping, cropping, brightness, and contrast adjustments. 
For all our experiments, we start from a randomly generated architecture to fairly compare with other SoTA methods. We first illustrated the search space in Tables \ref{tab:insert_space_image} for two search patterns, which is much larger than that proposed in EAS and PNAS, while we still have less GPU days for finding a good model. 

\noindent\textbf{Training details.} Our policy network uses LSTMs with 32 hidden units for network embedding, while LSTMs with 128 hidden units for \textit{scale} and \textit{insert} actions. It is trained with the Adam optimizer with a learning rate of 0.0006, with weights initialized uniformly in [-0.1, 0.1]. In total, 8 branches are constructed for training. 
Each generated neural networks are trained for 20 epochs with batch size (BZ) of 128 using Nesterov momentum with a learning rate (LR) ($l_{max}=0.05, l_{min}=0.001, T_{0}=10, T_{mul}=2$) following the cosine schedule\cite{SGDR}. With performance prediction ($p.p.$), we increase LR to 0.5 and BZ to 1024, and do early stop at 10 epochs for each model with parameter sharing. 
We use an episode step of 10 and 5 for layer-by-layer search and module search respectively, and select top 8 models from each episode to initialize the next. We run 15 episodes for searching the best model, which is retrained following PNAS, yielding our final performance.


\begin{table*}[t]
	\centering
    \fontsize{7}{8}\selectfont
    \caption{Comparison of EPNAS to automated architecture search literature for CIFAR-10. Note that ENAS GPU days are evaluated on our own server with author implementation, with the same system configuration with EPNAS's evaluation. M stands for million.}
    
	    \begin{tabular}{l|c|c|c|c}
	    \toprule[0.13em]
		Model & Parameters & Test error (\%) & Test error w cutout (\%) & GPU days \\\hline
 		AmoebaNet-B~\cite{Esteban.2018.REIC} & 2.8 M & 3.37 & - & 3150 \\ 
 		NASNet-A~\cite{Zoph2016} & 3.3 M & 3.41 & \textbf{2.65} & 1800 \\ 
 		Progressive NAS~\cite{Liu2017b} & 3.2M & 3.63 & - & 150 \\ 
 		EAS (DenseNet on C10+)~\cite{Han.2017.NT} & 10.7 M & 3.44 & - & 10 \\ 
		ENAS Macro Search~\cite{pham2018efficient} & 21.3 M & 4.23 & - & $1^{\ast}$ \\ 
		ENAS Micro Search~\cite{pham2018efficient} & 4.6 M & 3.54 & 2.89 & $1.5^{\ast}$ \\ 
		
		DARTS (first order)~\cite{liu2018darts} & 3.3 M & - & 4.23 & 1.5 \\ 
		DARTS (second order)~\cite{liu2018darts} & 3.3 M & - & 2.76 & 4 \\ \hline
		EPNAS Module Search & 4.0 M & 3.32 & 2.84 & 25 \\ 
		EPNAS Layer-by-Layer Search & 3.4 M & 3.87 & - & 16 \\ 
		EPNAS Module Search w prediction & 4.3 M & 3.2 & 2.79 & 8 \\ 
		EPNAS Layer-by-Layer Search w prediction & 7.8 M & \textbf{3.02} & - & 4 \\
		\toprule[0.13 em]
		\end{tabular}
	\label{tab:comparison_SOTA}
\end{table*}

\noindent\textbf{Ablation study.} Fig.~\ref{fig:na-accuracy} shows that EPNAS achieves the test accuracy up to 96.2\% with $p.p.$ and 95\% w/o $p.p.$ respectively after 6 GPU days, when started with a randomly generated model with a test accuracy of 91\%. Both layer-by-layer search and module search significantly outperform random search, but layer-by-layer search slightly outperforms module search as it enables more fine-grained search space. Cutout marginally improves searched accuracy by 0.5-0.6\%.  

\noindent\textbf{Quantitative comparison.} Tab.~\ref{tab:comparison_SOTA} compares EPNAS (with model size constraint) to other notable NAS approaches. All techniques yield similar accuracy (within $\sim$0.5\% difference) for comparable sizes. EPNAS with $p.p$ yields additional 1\% accuracy gain compared to EPNAS w/o $p.p$ as it enables searching more models within the same search time budget. EPNAS significantly outperforms AmoebaNet, NASNet, PNAS, and EAS in search time with larger search space. Specifically, comparing with PNAS, our module search w $p.p$ train roughly 600 models (15 episodes 8 M40 GPUs and 5 steps for each episode) to obtain a good model using module search (20 min per model with 10 epochs), and PNAS trained 1160 models with longer time to reach similar performances. 
EPNAS slightly outperforms both ENAS Macro Search and ENAS Micro Search in terms of model accuracy and model size, but is slightly worse in search time. 

\begin{table*}[!htpb]
	\centering
	\fontsize{6}{8}\selectfont
	\addtolength{\tabcolsep}{-4pt}
    \caption{EPNAS with multiple resource constraints vs. the SoTA models on CIFAR10. $sz$ means Model size (M). $c.i.$ means compute intensity (FLOPs/byte) and $c.c.$ means compute complexity (MFLOPs). 
    Compute intensity is not compute complexity divided by model size. It reflects how models reuse data without loading data from slow memory. The higher compute intensity is, the better. MFLOPs stands for mega floating-point operations per second.
    }
		\begin{tabular}{l|l|c|c|c|c}
	\toprule[0.13em]
	Model & Resource constraint  & Test error (\%) & Model size ($sz$) & \splitcell{Comp. intensity ($c.i.$) \\ (FLOPs/byte) } & \splitcell{Comp. complexity ($c.c.$) \\ (MFLOPs) } \\\hline\hline
	ResNet50~\cite{He_2017_ICCV} & -  & 6.97 & 0.86 M & 32 & 10 \\ 
 	DenseNet (L=40, k=12)~\cite{huang2017densely} & -  & 5.24 & 1.02 M& 43 & 21 \\ 
 	DenseNet-BC (k=24)~\cite{huang2017densely} & -& 3.62  & 15.3 M & 45 & 300\\
	ResNeXt-29,8x64d~\cite{xie2017aggregated} & -  & 3.65 & 34.4 M & 58 & 266\\ 
	MobileNetV2~\cite{mobilenetv2} & - & 5.8 & 1.9 M  & 10 & 10 \\
	DPPNet-PNAS~\cite{mobilenetv2} & - & 5.8 & 1.9 M  & 10 & 10 \\
	MNASNet~\cite{mobilenetv2} & - & 5.8 & 1.9 M  & 10 & 10 \\
	\hline
    EPNAS: Module Search &  $sz \leq$ 3M  & 3.98  & 2.2 M & 7.1 & 28\\ 
    EPNAS: Layer-by-Layer Search & $sz \leq$ 2M,~~ $c.i.\geq$ 80 & 4.31  & 1.7 M  & 97 & 26\\ 
    EPNAS: Layer-by-Layer Search &$c.i.\geq$100  ~~$c.c.\leq$ 200   & 2.95  & 29 M & 107 & 194\\ 
 	EPNAS: Layer-by-Layer Search & $sz \leq$ 8M,~~$c.i.\geq$ 80,~~$c.c.\leq$ 80    & 3.48 & 7.7 M & 92 & 72 \\
    EPNAS: Layer-by-Layer Search & $sz \leq$ 1M,~~$c.i.\geq$  30,~~$c.c.\leq$  15 & 5.95 & 0.88 M  & 31 & 13 \\ 
	\toprule[0.13em]
	\end{tabular}
    \vspace{-0.5\baselineskip}
	\label{tab:comparison_constrained}
\end{table*}

\noindent\textbf{EPNAS with resource constraints.} Table~\ref{tab:comparison_constrained} compares EPNAS and other SoTA models with resource constraints. 
In particular, EPNAS is able to find model under 10M parameters with 3.48\% test error with 92 FLOPs/Byte $c.i.$, and under 80 MFLOPs $c.c$. With tight model size and FLOPs constraints, EPNAS is able to find model of similar size compared to ResNet50 but is about 1\% more accurate. With relaxed model size, EPNAS finds SoTA model with 2.95\% test error and high $c.i.$ of 107 FLOPs/byte, which outperforms ENAS macro search. 


\begin{table}[t]
\fontsize{7}{9}\selectfont
    \caption{Comparison of model generalization to ImageNet. Model error rate ($\%$) is reported with Top-1 and Top-5 predictions. }
    \centering
    \begin{tabular}{l| l | c c  c c}
    \toprule[0.13em]
        Setting & Method & Top-1 & Top-5  & Params \\
        \hline
        \multirow{4}{*}{Mobile}  & DPP-Net-Panacea~\cite{dong2018dpp} & 25.98 & 8.21 &4.8M  \\
        & MnasNet-92~\cite{tan2018mnasnet} & 25.21 &7.95 & 4.4M\\
        & PNASNet-5~\cite{Liu2017b}  &  25.8 & 8.1 &5.1M \\
        & DARTS~\cite{liu2018darts} & 26.7 & 8.7 & 4.7M \\
        & EPNAS-mobile & \textbf{25.2} & \textbf{7.57} & 4.2M \\
        \hline
        \multirow{2}{*}{Large}  & DPP-Net-PNAS~\cite{dong2018dpp} & 24.16 & 7.13 & 77.16M \\
                                & EPNAS-large & 21.49 & 5.16 & 78.33M \\
 \toprule[0.13em]
    \end{tabular}
    \label{tab:imagenet}
   \vspace{-2.0\baselineskip}
\end{table}

\subsection{Results on ImageNet}
We also have experiments which transfer optimal models from EPNAS module search in Tab.~\ref{tab:comparison_SOTA} to ImageNet borrowing the stacking structure following PNAS~\cite{Liu2017b} with two settings:
\noindent1) \textit{Mobile}: limiting the model size to be less than 5M parameters,
\noindent2) \textit{Large}: without any model size constraint.
For a fair comparison, we set the input image size to 224 $\times$ 224 for both settings. The upper part in \tabref{tab:imagenet} shows the results of \textit{mobile} setting. We compare against other SoTA resource-aware NAS methods, DPPNet~\cite{dong2018dpp} and MNASNet~\cite{tan2018mnasnet}, and DARTS~\cite{liu2018darts}.
Our model based on the searched module yields better results with similar model size. 
For \textit{large} setting, our results is also significantly better than that reported in DPP-Net, while we omit that from PNAS since the input setting is different.

\vspace{-3mm}
\section{Conclusions}
\vspace{-2mm}
We propose efficient progressive NAS (EPNAS), a novel network transform policy with REINFORCE and an effective learning strategy. It supports large search space and can generalize well to NAS multiple resource constraints through a soft penalty function. 
EPNAS can achieve SoTA results for image recognition and KWS even with tight resource constraints.

{\small

}
\end{document}